\documentclass{article}

\usepackage{microtype}
\usepackage{graphicx}
\usepackage[inkscapearea=page]{svg}
\usepackage{adjustbox}
\usepackage{subfigure}
\usepackage{booktabs} 
\usepackage{xspace}
\usepackage{enumitem}

\usepackage{hyperref}

\usepackage[arxiv]{icml2024}

\usepackage{amsmath}
\usepackage{amssymb}
\usepackage{mathtools}
\usepackage{amsthm}

\usepackage{amsmath,amsfonts,bm}

\newcommand{\pn}[1]{}

\def\eqref#1{equation~\ref{#1}}

\def\floor#1{\lfloor #1 \rfloor}
\def\1{\bm{1}}

\DeclareMathAlphabet{\mathsfit}{\encodingdefault}{\sfdefault}{m}{sl}
\SetMathAlphabet{\mathsfit}{bold}{\encodingdefault}{\sfdefault}{bx}{n}

\newcommand{\R}{\mathbb{R}}
\newcommand{\Rd}{\mathbb{R}^d}

\newcommand{\norm}[1]{\left\|#1\right\|}

\newcommand{\baseline}{\textrm{Baseline}}
\newcommand{\cA}{\mathcal{A}}
\newcommand{\x}{\overrightarrow{\mathbf{x}}}
\newcommand{\y}{\overrightarrow{\mathbf{y}}}
\newcommand{\z}{\overrightarrow{\mathbf{z}}}

\newcommand{\uvec}{\overrightarrow{\mathbf{u}}}
\newcommand{\vvec}{\overrightarrow{\mathbf{v}}}
\newcommand{\U}{{\mathbf{U}}}
\newcommand{\W}{{\mathbf{W}}}
\newcommand{\Z}{{\mathbf{Z}}}
\newcommand{\Zapprox}{{\mathbf{Z}_{\textrm{approx}}}}
\newcommand{\Wapprox}{{\mathbf{W}_\textrm{approx}}}

\newcommand{\Stilde}{\widetilde{S}}

\newcommand{\sm}[1]{\textrm{Softmax}\left(#1\right)}

\newcommand{\smk}[1]{\textrm{Softmax}_k\left(#1\right)}

\newcommand{\ff}[1]{\textrm{FF}\left(#1\right)}
\newcommand{\ffk}[1]{\textrm{FF}_k\left(#1\right)}
\newcommand{\fftildek}[2]{\widetilde{\textrm{FF}}_{#2}\left(#1\right)}
\newcommand{\iprod}[2]{\langle #1, #2 \rangle}
\newcommand{\topk}[1]{\topkinp{#1}{k}\xspace}
\newcommand{\topkind}[1]{\topind{#1}{k}\xspace}
\newcommand{\topkinp}[2]{\textrm{Top}_{#2}\left(#1\right)\xspace}
\newcommand{\topind}[2]{\textrm{TopInd}_{#2}\left(#1\right)\xspace}
\newcommand{\grouptopind}[2]{\textrm{GroupTopInd}_{#2}\left(#1\right)\xspace}
\newcommand{\dmodel}{d_{\textrm{m}}\xspace}
\newcommand{\bfsixteen}{\textrm{bf16}\xspace}

\newcommand{\intfour}{\textrm{int4}\xspace}

\newcommand{\Mbase}{\mathcal{M}_{\textrm{base}}\xspace}
\newcommand{\Mffn}{\mathcal{M}_{\textrm{ffn}}\xspace}
\newcommand{\Msm}{\mathcal{M}_{\textrm{sm}}\xspace}
\newcommand{\Mfull}{\mathcal{M}_{\textrm{full}}\xspace}
\newcommand{\herd}{\textrm{HiRE}\xspace}
\newcommand{\herdlr}{\herd-\textrm{LR}\xspace}
\newcommand{\herdq}{\herd-\textrm{Q}\xspace}
\newcommand{\herdlrq}{\herd-\textrm{LRQ}\xspace}
\newcommand{\herdsm}{\herd-\textrm{Softmax}\xspace}
\newcommand{\herdffn}{\herd-\textrm{FFN}\xspace}
\newcommand{\paramsize}[1]{\textrm{PS}\left(#1\right)\xspace}
\newcommand{\abs}[1]{\left|#1\right|\xspace}
\newcommand{\distop}{\textrm{DA-TOP-}k\xspace}

\usepackage{hyperref}
\usepackage{url}
\usepackage{array}
\usepackage{graphicx}
\usepackage{multirow}
\usepackage{makecell}

\usepackage[capitalize,noabbrev]{cleveref}

\theoremstyle{plain}

\theoremstyle{definition}

\theoremstyle{remark}

\usepackage[textsize=tiny]{todonotes}

\icmltitlerunning{\herd: High Recall Approximate Top-$k$ Estimation for Efficient LLM Inference}

\begin{document}

\twocolumn[
\icmltitle{\herd: High Recall Approximate Top-$k$ Estimation for Efficient LLM Inference}



\icmlsetsymbol{equal}{*}
\icmlsetsymbol{equalsenior}{\textdagger}

\begin{icmlauthorlist}
\icmlauthor{Yashas Samaga B L}{equal,gri}
\icmlauthor{Varun Yerram}{equal,gri}
\icmlauthor{Chong You}{grnyc}
\icmlauthor{Srinadh Bhojanapalli}{grnyc}
\icmlauthor{Sanjiv Kumar}{grnyc}
\icmlauthor{Prateek Jain}{equalsenior,gri}
\icmlauthor{Praneeth Netrapalli}{equalsenior,gri}
\end{icmlauthorlist}

\icmlaffiliation{gri}{Google Research, India}
\icmlaffiliation{grnyc}{Google Research, New York City}

\icmlcorrespondingauthor{Yashas Samaga}{syashas@google.com}
\icmlcorrespondingauthor{Varun Yerram}{vyerram@google.com}
\icmlcorrespondingauthor{Praneeth Netrapalli}{pnetrapalli@google.com}

\icmlkeywords{Machine Learning, ICML}

\vskip 0.3in
]



\printAffiliationsAndNotice{\icmlEqualContribution} 

\begin{abstract}

Autoregressive decoding with generative Large Language Models (LLMs) on accelerators (GPUs/TPUs) is often memory-bound where most of the time is spent on transferring model parameters from high bandwidth memory (HBM) to cache. 
On the other hand, recent works show that LLMs can maintain quality with significant sparsity/redundancy in the feedforward (FFN) layers by appropriately training the model to operate on a top-$k$ fraction of rows/columns (where $k \approx 0.05$), there by suggesting a way to reduce the transfer of model parameters, and hence latency. However, exploiting this sparsity for improving latency is hindered by the fact that
identifying top rows/columns is data-dependent and is usually performed using full matrix operations, severely limiting potential gains. To address these issues, we introduce \herd (\textbf{Hi}gh \textbf{R}ecall {Ap}proximate Top-$k$ \textbf{E}stimation). \herd comprises of two novel components:  (i) a compression scheme to cheaply predict top-$k$ rows/columns with high recall, followed by full computation restricted to the predicted subset, and (ii) $\distop$: an efficient multi-device approximate top-k operator. 
We demonstrate that on a one billion parameter model, \herd applied to both the softmax as well as feedforward layers, achieves almost matching pretraining and downstream accuracy, and speeds up inference latency by $1.47\times$ on a single TPUv5e device~\cite{tpuv5e}.
\end{abstract}

\section{Introduction}
Deploying large language models (LLMs) is almost prohibitively expensive due to cost of running inference on accelerators and also has significant environmental costs. For example, \cite{patterson2021carbon} attribute $80-90\%$ of the total carbon emissions during the life cycle of an LLM to running inference. 

Latency/cost of generative LLMs is dominated by the autoregressive next-token generation which in-turn is memory-bound on standard accelerators (GPU/TPUs) due to shuttling large matrices from high-bandwidth memory (HBM) to the accelerator cache. Several recent approaches seek to address this challenge by using block parallel decoding~\cite{stern2018blockwise}, speculative decoding~\cite{leviathan2023fast}, mixture of experts~\cite{du2022glam}, more efficient architectures~\cite{gu2023mamba} as well as by using classical methods such as distillation~\cite{liang2020mixkd}.

Despite these techniques, generative LLM inference is still primarily memory-bound, indicating significant room for latency/cost reduction. Furthermore, for typical input length (e.g. $\lessapprox 2000$), and even for models with up to one billion parameters, softmax and feedforwad (FFN) layers account for $\geq 90\%$ of the latency. So in this work, we primarily focus on these two components\footnote{while our ideas are applicable to the attention layer as well, we leave a more detailed investigation of that to future work.}. Multiple recent works show that LLMs have inherent ``sparsity" or redundancy which seem to actually grow with larger models. For example, \cite{li2022lazy,mirzadeh2023relu} show that the feedforward (FFN) layers activations -- with ReLU based activation functions -- are very sparse. More concretely, for a $1$-hidden layer FFN: $\x_{\textrm{out}} = \sum_{j=1}^m \phi(\iprod{\uvec_j}{\x})\vvec_j$ with input $\x_{\textrm{in}}$, output $\x_{\textrm{out}}$, $\phi(\cdot)$ a ReLU based activation function and $\uvec_j$ and $\vvec_j$ being the first and second layer weights of the FFN respectively, for any input $\x_\textrm{in}$, only a small fraction of weight vectors $\uvec_j$ have $\phi(\iprod{\uvec_j}{\x}) \neq 0$. In fact, the sparsity level can be further enhanced ($\lessapprox 5\%$ non-zeros) by using explicit top-$k$ procedure in the FFN layer computation during training. Similarly, softmax layers usually need to focus only on tokens corresponding to the highest magnitude outputs (logits).

This raises the following question that we seek to answer: {\em can we compute top elements of softmax output and/or FFN activations efficiently (on accelerators) and accurately?} 

{\bf Challenges with top-k computation in softmax/FFN}: Firstly, the top-k elements can be input dependent, so the challenge is to estimate top elements without fully evaluating the softmax/FFN layer. Second, even after estimating the top elements' coordinates, it is challenging to gather the top elements efficiently on standard accelerators (GPUs/TPUs).  

{\bf Contributions}: In this work, we propose ideas to overcome these challenges and show that they substantially improve the inference latency of autoregressive LLMs.

\textbf{Estimating top-$k$ elements efficiently}: We propose using an \textbf{approximate} but \textbf{high recall} procedure to estimate the indices of top-$k$ outputs of FFN/softmax layers. In particular, we employ a smaller dimensional/low rank projection as well as aggressive quantization for doing this approximate estimation, but the framework is more general. While there are a few recent works~\cite{liu2023deja,csordas2023approximating,alizadeh2023llm} which employ an approximate top-$k$ estimation procedure, they suffer a loss in performance due to the approximation. One of our key ideas is to identify that if this approximate procedure has \emph{high recall}, then following it by exact computation on the predicted set can regain the lost performance. We call this approach \herd, standing for \textbf{Hi}gh \textbf{R}ecall {Ap}proximate Top-$k$ \textbf{E}stimation.

{\bf Distributed approximate top-$k$ (\distop)}: Due to the size of LLMs, even during inference, the models are sharded onto multiple devices (say $\ell$ devices), and similarly computations (and outputs/activations) in each of the softmax/FFN layers are also distributed across the devices. In this context, a vanilla application of the top-$k$ operator turns out to be very expensive since parameters stored on all the different devices are collected into a central location before performing the top-$k$ operation. To overcome this, we propose a \emph{distributed, approximate top-$k$} operator which performs the top-$\frac{k}{\ell}$ operation on each of the $\ell$ devices, and then uses a union of these to approximate the overall top-$k$ operator.

{\bf \herdsm}: In the softmax layer, we are interested only in the top few ($k\lessapprox 32$) outputs from a very large output space with hundreds of thousands of tokens. On a model with about one billion parameters, our implementation of \herd gives an end-to-end latency improvement of about $1.22\textrm{x}$ without any quality drop.

{\bf \herdffn}: While LLMs trained with a top-$k$ operation have relatively sparse activations, $k \gtrapprox 5\%$ is usually required to ensure that there is no loss in accuracy (in contrast to the softmax layer, where the effective sparsity is $\lessapprox 0.3\%$). Now, the amount of time taken to gather the relevant top-$k$ columns of the weight matrix from HBM to cache turns out to be substantially more than that required to bring an equivalently sized dense matrix. To overcome this hardware limitation, we propose a group sparse top-$k$ operation, where columns of the weight matrix are grouped into groups of small sizes ($8$ in our experiments) which significantly improves the efficiency of memory transfers. Second, motivated by the observation that there is substantial overlap in non-zero activations across tokens, we propose two approaches to further enhance sparsity levels. For exploiting {static overlap} (i.e., activations that are common to most tokens), we propose to use a hybrid FFN layer which combines a standard/dense (but small) FFN, together with a sparse/top-$k$ FFN layer. For exploiting {dynamic overlap} (i.e., activations 
that are common to related responses), while doing inference with a larger batch size or generating multiple parallel responses for the same query, we propose to select sparsity depending on the size of \emph{union of non-zero activation indices} for each of the tokens, instead of using a fixed level of sparsity. On the same model as above, \herdffn gives an end-to-end latency improvement of about $1.16\textrm{x}$, again without any quality drop.

{\bf Putting together}: A combination of all the ideas discussed above give an end-to-end latency improvement of about $1.47\textrm{x}$ compared to the fully dense model.

We note that concurrent to our work, ~\cite{alizadeh2023llm} also proposes an efficient, approximate procedure to estimate non-zero activations. While this approach is similar to our ``estimating top activations cheaply" contribution, we note that emphasis of the two works is quite different. In particular, we demonstrate that along with estimation of top activations, we also need to adjust thresholds for high recall,  implement \distop style operator, and promote group sparsity+overlap to provide compelling end-to-end latency reduction on accelerators. 

{\bf Outline}: In Section~\ref{sec:related}, we describe some closely related works. In Section~\ref{sec:idea}, we explain the key ideas. In Section~\ref{sec:exp} we present the experiment setting and our main results, and provide additional results in appendix. 

{\bf Notation}: We use non-bold letters $a, A$ etc. for scalars, $\overrightarrow{\mathbf{a}}, \overrightarrow{\mathbf{b}}$ etc. for vectors and $\mathbf{A}, \mathbf{B}$ etc. for matrices. Given a set $S$ of coordinates, we also use $\x\vert_S$ to denote the restriction of $\x$ to the coordinates in $S$.
\section{Related Work}\label{sec:related}
{\bf Sparsity of activations}: As mentioned earlier, recent works~\cite{li2022lazy} have demonstrated that large language models have highly sparse activation patterns with ReLU based activation functions.~\cite{mirzadeh2023relu} show that using ReLU activation indeed achieves comparable performance to and argue for reinstating the ReLU activation function instead of more popular alternatives such as GeLU. There have been several recent attempts to exploit activation sparsity for faster inference such as~\cite{liu2023deja,grimaldi2023accelerating,piorczynski2023exploiting,csordas2023approximating,alizadeh2023llm}, but all of them suffer some loss in quality since the activation pattern cannot be predicted exactly. In contrast, one of our key insights is that estimating activation sparsity with \emph{high recall} can ensure matching quality, since exact computation can be performed on all of the activations that are estimated to be non-zero, and even if some of them later turn out to be zero, it doesn't affect the actual output. We also note that we introduce several new ideas such as group sparsity with very small group sizes, using a common dense path, exploiting activation overlap across tokens, and distributed gathering of weights, which are critical to improving efficiency without loss in accuracy and in extending this approach to larger batch sizes, larger number of samples and larger models with model partitioning.

{\bf Sparse attention}: There have also been several approaches to \emph{enforce} and exploit sparsity in the attention layers~\cite{wang2021spatten,madaan2022treeformer}. In particular, they directly approximate full all-to-all attention with sparse attention and we believe that similar to what we do for activation sparsity in this paper, estimating a high recall set of non-zero attention weights and performing exact attention computation on them can ensure that there is no loss in quality with sparse attention.

{\bf Structured sparsity and hardware support}: From the hardware/systems point of view, there have been several works developing efficient sparse operations on GPUs and TPUs such as~\cite{gale2020sparse}. Since unstructured sparsity is not supported with efficient execution on GPUs, there have been works that try either different forms of structured sparsity such as $1$ in $4$ sparsity~\cite{jaszczur2021sparse}, group sparsity with large group sizes~\cite{dong2023towards} or explore the efficiency of sparse transformers on CPUs~\cite{song2023powerinfer,zeng2023lookupffn}. In this work, we show that while fully unstructured sparsity is not efficiently supported on TPUs, surprisingly, group structured sparsity with even small group sizes like $8$, is efficiently supported on TPUs (see Figure~\ref{fig:group-sparsity-efficiency}), while not losing in terms of pretraining or downstream quality.

{\bf Mixture of experts (MoE)}: MoE~\cite{du2022glam} can be thought of as an extreme version of structured sparsity, where the group size is very large, and have been explored both for training~\cite{li2022lazy} as well as inference~\cite{yi2023edgemoe} efficiency. However, MoE models are usually harder to train, since one needs to both learn a gating mechanism, as well as ensure that tokens are roughly equally distributed across experts. In contrast, small group sizes (or experts) are much easier to train, particularly with the Top-$k$ operation.

{\bf Complementary approaches for inference efficiency}: There have been several complementary approaches for speeding up LLM inference such as model compression~\cite{han2015deep,jaiswal2023compressing,xia2023flash}, quantization~\cite{li2023norm,ahmadian2023intriguing,lin2023awq,zhao2023atom}, speculative decoding~\cite{leviathan2023fast,he2023rest}, early exit and parallel decoding~\cite{bae2023fast}, structured matrices in FFN layers~\cite{baykal2023alternating}, other systems aspects such as effective partitioning of the model~\cite{pope2023efficiently}, communication protocols across multiple devices~\cite{aminabadi2022deepspeed,sheng2023flexgen} etc.
\section{Main Ideas}\label{sec:idea}
In this section, we will introduce the precise setting and present the main ideas behind \herd more rigorously.
\subsection{Problem Setup}
A transformer begins with an \emph{embedding} layer which produces the representation for the input tokens, then processes them through a sequence of alternating \emph{attention} and \emph{feed forward} layers, and finally ends with a \emph{softmax} layer that produces the output probabilities~\cite{vaswani2017attention}. In this work, we will be primarily concerned with the \emph{softmax} and \emph{feed forward (FFN)} layers, which we now describe.

The \textbf{softmax} layer takes an input $\x\in \Rd$ and outputs:
\begin{align}\label{eqn:softmax}
    \sm{\x} := \frac{\exp(\W\x)}{\norm{\exp(\W\x)}_1} \in \R^c,
\end{align}
where $\W \in \R^{d \times c}$, $d$ is called the model dimension and $c$ is the number of output classes. In almost all of the applications, we only care about accurately estimating the conditional distribution on the top-$k$ output probabilities for some $k$ i.e.,
\begin{align}\label{eqn:softmax-top-k}
    \smk{\x} := \frac{\topk{\sm{\x}}}{\norm{\topk{\sm{\x}}}_1} \in \R^c,
\end{align}
where $\topk{\cdot}$ operation takes a vector as input, retains the values of the top-$k$ entries and sets the remaining to zero. In particular, we only care about estimating the location and values of the top-$k$ entries of $\sm{\x}$.

The \textbf{FFN} layer takes an input $\x \in \Rd$ and outputs:
\begin{align} \label{eqn:feedforward}
    \ff{\x} := \sum_{j=1}^m \phi(\iprod{\uvec_j}{\x}) \vvec_j,
\end{align}
where $\phi(\cdot)$ is an activation function, $m$ is the number of hidden units and $\uvec_j \in \Rd$ and $\vvec_j \in \Rd$ are the first and second layer weights respectively. We refer to~\eqref{eqn:feedforward} as the standard or dense FFN layer.
It turns out that for specific activation functions like (powers of) ReLU, for any given $x$, the number of $j$ for which the activations i.e., $\phi(\iprod{\uvec_j}{\x})$ is non-zero in a trained LLM is small ($\lessapprox 10\%$)~\cite{li2022lazy,mirzadeh2023relu}. If we explicitly do a top-$k$ step in the feedforward evaluation i.e.,
\begin{align}\label{eqn:ff-topk}
    \ffk{\x} := \sum_{j\in S(\x)} \phi(\iprod{\uvec_j}{\x}) \vvec_j,
\end{align}
where $S(\x) := \topkind{\phi(\mathbf{U}^\top{\x})}$, where $\mathbf{U}$ is the matrix whose $j^\textrm{th}$ column is $\uvec_j$ and $\topkind{\z}$ denotes the set of indices where the top-$k$ entries occur in $\z$, then the number of non-zeros reduces even further without drop in quality for $k\approx 5\%$~\citep{li2022lazy}. We refer to~\eqref{eqn:ff-topk} as the sparse or top-$k$ FFN layer.

We now abstract out the key component that can speed up both top-$k$ based softmax~\eqref{eqn:softmax-top-k} as well as top-$k$ based FFN~\eqref{eqn:ff-topk} as follows. Given a matrix $\Z \in \R^{d \times \ell}$ and a vector $\x \in \R^d$, we wish to compute:
\begin{align}\label{eqn:general-topk}
    S = \{(i,\phi(\iprod{\z_i}{\x})): i \in \topkind{\phi(\Z^\top{\x})}\},
\end{align}
where $\z_i$ is the $i^{\textrm{th}}$ column of $\Z$ and $\phi(\cdot)$ is any given activation function (could be identity for softmax~\eqref{eqn:softmax-top-k}), and $d \ll \ell$. Our goal is to design an efficient mechanism to compute $S$ in~\eqref{eqn:general-topk} compared to the baseline approach of computing $\phi(\Z^\top{\x})$ followed by a top-$k$ operation.

\vspace{-2mm}\paragraph{Theoretical proxy for quantifying efficiency}: Autoregressive decoding for small batch sizes is memory bound i.e., the time taken for transferring model parameters across different hierarchies of memory (e.g., HBM/RAM to cache) of the accelerator device (GPU/TPU) is the largest component of the total inference latency~\cite{leviathan2023fast}.
To quantify this intuition, and get a sense of how much we are improving inference latency, \emph{only in the current section}, we use the number of effective parameters (measured in terms of bytes taken to store them) used by a given algorithm $\cA$ for the computation in~\eqref{eqn:general-topk} as a proxy for its latency, and refer to it as $\paramsize{\cA}$. For example, the $\baseline$ which first computes $\phi(\Z^\top{\x})$, followed by a top-$k$ operation has $\paramsize{\baseline} = 2d\ell$ since $\Z$ is a $d\times \ell$ matrix, which is stored in $16$-bit floating point ($\bfsixteen$) format.

\subsection{\herd: Overall Approach}
In this section, we present the key ideas behind \herd, along with its application to the softmax and FFN layers.
\subsubsection{Solving~\eqref{eqn:general-topk} accurately and efficiently}
\begin{algorithm}
  \caption{Pseudocode for \herd}
  \label{alg:herd}
  \begin{algorithmic}[1]
    \INPUT $\x, \Z, \Zapprox, k, k'$
    \STATE $S' \leftarrow \topind{\phi\left(\Zapprox^\top {\x}\right)}{k'}$ \COMMENT{Top-$k'$ indices of $\phi\left(\Zapprox^\top {\x}\right)$}
    \STATE $\y \leftarrow {\topk{\Z\vert_{S'}^\top \x}}$
    \OUTPUT $\y$
  \end{algorithmic}
\end{algorithm}
\begin{figure}[t]
    \centering
    \includegraphics[scale=0.3]{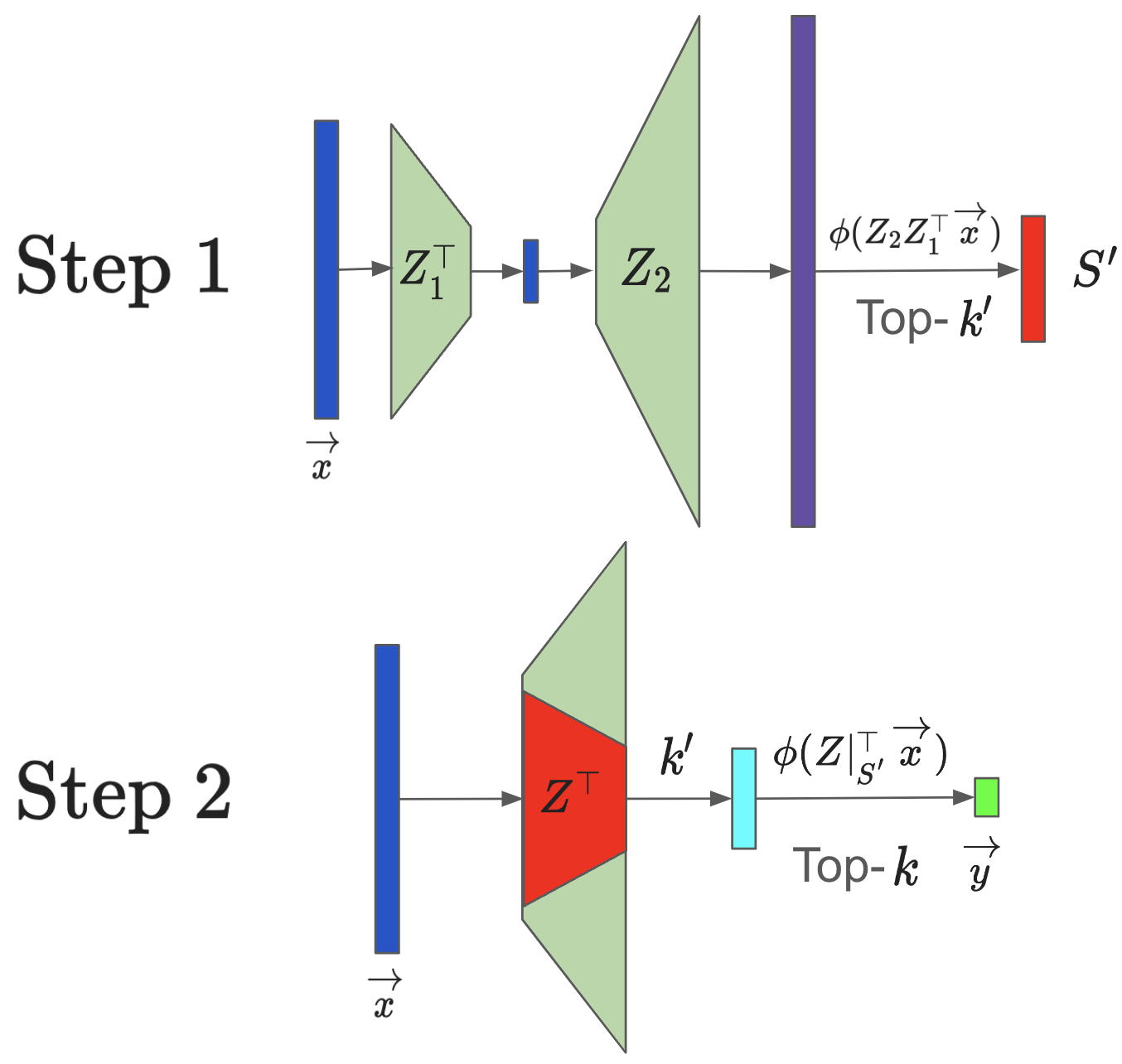}\vspace*{-5pt}
    \caption{\herd schematic: To compute the top-$k$ elements of $\phi(\Z^\top \x)$, we first compute an approximate top-$k'$ index set $S'$ by using a low rank approximation $\Zapprox = \Z_1 \Z_2^\top$. We then compute $\phi(\Z\vert_{S'}^\top \x)$ for $\Z$ restricted to $S'$ and then perform top-$k$ operation on that vector.}
    \label{fig:herd-schematic}
\end{figure}
The key idea behind \herd is that given an approximate, and smaller version of the matrix $Z$, denoted by $\Zapprox$, we first approximate $\phi(\Z^\top \x)$ with $\phi(\Zapprox^\top \x)$, use it to compute the set of top-$k'$ elements $S'$ for some $k' > k$, compute $\phi(\Z\vert_{S'}^\top \x)$ for only indices restricted to $S'$, and then perform the top-$k$ operation on the resulting vector. A pseudocode of \herd is presented in Algorithm~\ref{alg:herd} and a schematic diagram is presented in Figure~\ref{fig:herd-schematic}. It is easy to see that as long as $\topind{\phi(\Z^\top \x)}{k} \subseteq S'$, we have that $\topk{\phi(\Z^\top \x)} = \topk{\phi(\Z\vert_{S'}^\top \x)}$. If the size of $\Zapprox$ is chosen to be substantially smaller than that of $\Z$, and if $k' \ll \ell$, then \herd executes much faster than the naive version of implementing~\eqref{eqn:general-topk}. In this paper, we consider two ways of choosing $\Zapprox$.
\vspace{-5mm}\paragraph{\herdlr}: In this case, we choose $\Zapprox$ to be a low rank matrix factorized as $\Z_1 * \Z_2^\top$, where $\Z_1 \in \R^{d\times r}$ and $\Z_2 \in \R^{\ell \times r}$ for some $r \ll d$. In this case, $\paramsize{\herdlr} = 2(dr + r \ell + dk')$, which will be much smaller than $2d\ell$ since $r \ll d \ll \ell$ and $k' \ll \ell$. Note that this requires us to train $\Zapprox$ either in an end to end manner, or by performing low rank decomposition on $\Z$. However, there is a high potential for latency gains here since $r$ could potentially be chosen to be very small.
\vspace{-5mm}\paragraph{\herdq}: In this case, we choose $\Zapprox$ to be a aggressively quantized version of $\Z$ for example in $4$-bit integer format (\intfour). In this case, $\paramsize{\herdq} = \frac{d\ell}{2} + 2dk'$, which will be much smaller than $2d\ell$ since $k' \ll \ell$. Note that this does not require any training since we can just apply a standard quantization routine to $\Z$ to obtain $\Zapprox$. However, the potential gains are limited due to inherent limits on quantization.
Note that we can also combine both \herdlr and \herdq to obtain further inference latency improvements.

\subsubsection{$\distop$: Multi-device serving with approximate distributed top-$k$ operation}
\begin{algorithm}
  \caption{Pseudocode for \herd with \distop}
  \label{alg:distop}
  \begin{algorithmic}[1]
    \INPUT $\x, \Z, \Zapprox, k, k'$ with $\Z$ and $\Zapprox$ distributed across $s$ machines.
    \STATE On machine $i\in [s]$: $S'_i \leftarrow \topind{\phi\left(\Zapprox_i^\top {\x}\right)}{\frac{k'}{s}}$ \COMMENT{Top-$\frac{k'}{s}$ indices of $\phi\left(\Zapprox_i^\top {\x}\right)$ on machine $i$.}
    \STATE $\y_i \leftarrow {\topkinp{\Z\vert_{S'}^\top \x}{\frac{k}{s}}}$
    \STATE $y \leftarrow \textrm{Concat}(y_i: i \in [s])$.
    \OUTPUT $\y$
  \end{algorithmic}
\end{algorithm}
Very large models do not fit on the RAM of a single GPU/TPU, so for serving such models, the parameters of the model are usually distributed on a cluster of multiple devices. In this case, using \herd (Algorithm~\ref{alg:herd}) as is leads to large costs in communication. To mitigate the communication costs, we modify \herd to use \emph{distributed, approximate top-$k$ computation}. The pseudocode for the resulting algorithm is given in Algorithm~\ref{alg:distop}.
We will now present the application of Algorithms~\ref{alg:herd} and~\ref{alg:distop} to make softmax and FFN layers more efficient.
\subsection{\herd-SoftMax: Efficient Softmax using \herd}\label{sec:herd-softmax}
Since we usually care only about the top-$k$ logits of the softmax layer~\eqref{eqn:ff-topk}, we can directly apply Algorithms~\ref{alg:herd} and~\ref{alg:distop} to solve~\eqref{eqn:softmax-top-k} more efficiently. We note that while obtaining $\Wapprox$, an approximate version of $\W$, is straightforward with quantization, obtaining a low rank approximation of $\Wapprox$ is also relatively cheap since we can distill the inputs and outputs of $\W$ to obtain $\Wapprox$.
\subsection{\herd-FFN: Efficient FFN using \herd}\label{sec:herd-ffn}
Recalling~\eqref{eqn:ff-topk}, we note that we can compute $S(\x):=\topkind{\phi(\mathbf{U}^\top{\x})}$ using \herd.

While \herd-FFN is indeed theoretically more efficient compared to a naive computation of~\eqref{eqn:ff-topk}, it turns out that for the relative values of $k'$ and $m$ that are needed to ensure that there is no accuracy drop ($\frac{k'}{m} \approx 0.05$), the transfer of parameters across different hierarchies of memory is too inefficient to give any gains\footnote{In contrast, the corresponding factor in the softmax layer is much smaller, around $10^{-4}$, which gives us inference latency improvements, despite this inefficiency in memory transfer.}. Our key insight is that group sparse structure substantially improves the efficiency of parameter transfer across different hierarchies of memory. For example, Figure~\ref{fig:group-sparsity-efficiency} in Appendix~\ref{app:additional-results} shows that the efficiency of the transfer operation of columns of a matrix across different hierarchies of memory improves substantially when we transfer \emph{groups} of adjacent columns instead of single columns.
More concretely, we first divide the entire set of $m$ hidden units into $m/g$ groups with $g$ hidden units each (in all our experiments, we use $g=8$). Given an approximate group activation computation procedure $\Phi:\R^d \rightarrow \R^{{\frac{m}{g}}}$, i.e., $\Phi(\x) \approx \left(\sum_{k=0}^{g-1} \abs{\phi(\iprod{\uvec_{jg+k}}{\x})}: j\in[m/g]\right)$ let us denote $S'_g := \topind{\Phi(\x)}{k'}$, and use:
\begin{align}\label{eqn:herd-groupsparse}
    \fftildek{\x}{g,\frac{k'}{g}} := \sum_{j \in \Stilde_g} \sum_{\ell=0}^g \phi(\iprod{\uvec_{g*j + \ell}}{\x}) \vvec_{g*j + \ell},
\end{align}
where $\Stilde_g \subseteq [m/g]$ is a subset of groups selected as $\Stilde_g := \grouptopind{\phi(\U \vert_{{S'}_g}^\top \x)}{g,k/g}$, where ${S'}_g$ is an estimate of the groups of neurons that will actually be activated. We can again use Algorithm~\ref{alg:herd} (resp. Algorithm~\ref{alg:distop}) to compute $\Stilde_g$ in the single (resp. multiple) device settings.
\subsubsection{Enhancing sparsity further by exploiting static and dynamic activation overlap}\label{sec:overlap}
Motivated by the observation that there is substantial overlap in non-zero activations across tokens, we also propose the following approaches to exploit static and dynamic activation overlap.
\vspace{-4mm}\paragraph{Static overlap}: The first idea is to augment the sparse FFN layer with a \emph{dense common path} of neurons that are activated for all tokens, so that the number of non-zeros in the sparse part can be reduced further. More concretely, the feedforward layer comprises of two sets of $m_1$ and $m_2$ hidden units respectively, where the first set of neurons is used in a dense manner~\eqref{eqn:feedforward}, while the second set is used in a group sparse manner~\eqref{eqn:herd-groupsparse}. We have:
\begin{align}
    & \fftildek{\x}{\textrm{CommonPath}} = \sum_{j=1}^{m_1} \phi(\iprod{\uvec_j^{d}}{\x})\vvec_j^{d} \nonumber \\ &\quad + \sum_{j \in \Stilde_g(\x)} \sum_{\ell=0}^g \phi(\iprod{\uvec_{g*j + \ell}^{gs}}{\x}) \vvec_{g*j + \ell}^{gs}, \label{eqn:commonpath-ffn}
\end{align}
where $\uvec^d, \vvec^d$ denote the first set of hidden units used in a dense manner, while $\uvec^{gs}, \vvec^{gs}$ denote the second set of neurons used in a sparse manner.
\vspace{-4mm}\paragraph{Dynamic overlap}: In order to further improve efficiency while producing a larger number of samples for the same query (e.g., to rank these responses and output the best one~\cite{mudgal2023controlled}), we first compute $\Stilde_g(\x_1), \cdots, \Stilde_g(\x_s)$ for the latest token from each of the $s$ samples and use their union $\cup_{u\in [s]} \Stilde_g(\x_u)$ on line $1$ of Algorithm~\ref{alg:herd} and~\ref{alg:distop}.

\begin{table*}[t!]
\caption{Evaluation of \herdsm with the baseline dense model $\Mbase$ on a single TPUv5e device. For \herdsm, we use three different kinds of approximation: \herdlr with low rank approximation, \herdq with \intfour quantization and then finally \herdlrq with both low rank and \intfour quantization. We notice similar speedups with both \herdq and \herdlr as both of them reduce the size of parameters that are used for computation by $25\%$.}\vspace*{-5pt}
\label{tab:softmax-main-table}
\begin{center}
\begin{tabular}{|c|c|c|c|c|c|}
 \hline
  \multicolumn{3}{|c|}{} & \multicolumn{3}{|c|}{$\Msm$} \\
\hline
 \multicolumn{2}{|c|}{} & \makecell{$\Mbase$} & \makecell{HiRE-LR \\ ($r=25\%$, $k'=384$)} & \makecell{HiRE-Q \\ (int4, $k'=128$)} & \makecell{HiRE-LR \\ ($r=25\%$, $k'=384$) \\+ HiRE-Q  (int4)} \\ 
 \hline
 \multirow{2}{*}{\makecell{Pre-training \\ Performance}} & Top1 Accuracy & 57.15\% & 57.07\% & 57.12\% & 57.07\% \\
 \cline{2-6}
  & Top32 Intersection & 32.0 & 29.26 & 31.48 & 29.03 \\
 \hline
 \hline
 \multirow{4}{*}{\makecell{Downstream \\ Performance}} & Machine Translation & 47.92 & 47.73 & 47.9 & 47.77 \\ 
 \cline{2-6}
 & SuperGLUE Benchmark & 62.0 & 61.39 & 61.56 & 61.02 \\ 
 \cline{2-6}
 & Question Answering & 29.65 & 29.58 & 29.54 & 29.58 \\ 
 \cline{2-6}
 & Discriminative Tasks & 51.69 & 50.51 & 50.92 & 50.40 \\ 
 \hline
 \hline
 \multicolumn{2}{|c|}{\emph{Speedup}} & 1.0x & 1.16x & 1.16x & 1.22x \\
 \hline
\end{tabular}
\end{center}
\end{table*}

\section{Experiments}\label{sec:exp}
In this section, we present experimental evaluation of HERD and demonstrate its efficacy in reducing inference latency.
\subsection{Training details}\label{sec:exp-training}
We mostly follow the training setup described in~\cite{anil2023palm} including the training dataset and optimizer. Most  experiments were conducted on a model with about $1$ billion parameters -- we denote it by $1B$ model. $\dmodel$ denotes its model dimensions, i.e., dimension of representations.

\textbf{\herdffn}: After  pretraining $1B$ for a large number of steps ($\mathcal{M}$), we continue to pretrain it for a similar number of additional steps with group sparse top-$k$ operation (\eqref{eqn:herd-groupsparse}) on odd FFN layers to obtain $\Mffn$ i.e., if the model has $\ell$ layers, then we modify only the $1^\textrm{st}, 3^\textrm{rd},...,(2*\floor{\frac{\ell}{2}}-1)^{\textrm{th}}$ layers to be group sparse. For fair comparison we continue to pretrain $\mathcal{M}$ for the same number of additional steps as $\Mffn$, but without the top-$k$ operation to obtain the baseline model $\Mbase$. While the exact computations are all performed in $\bfsixteen$ (i.e., $16$ bit floating point), the approximate computations in \herd are performed in $\intfour$.

\textbf{\herdsm}: We apply \herdsm to $\Mbase$ or $\Mffn$ with different choices for approximate computation.
    \begin{enumerate}[leftmargin=*,noitemsep,nolistsep]
        \item For approximate computation with \emph{low rank}, we train $\Wapprox$ using cross entropy (CE) loss with respect to ground truth labels, with rank $r$ chosen to be $r=\frac{\dmodel}{4}$.
        \item For approximate computation with \emph{quantization}, we perform the softmax computation in $\intfour$.
    \end{enumerate}
    We also present results combining both low rank and quantization approximation. We refer to the model with \herd on both Softmax and FFN layers as $\Mfull$, while \herdsm applied to $\Mbase$ is referred to as $\Msm$.

\begin{table*}[t!]
\caption{Evaluation of \herdffn, as well as a combination of \herdsm + \herdffn with the baseline dense model $\Mbase$ on a single TPUv5e device. For \herdffn, we use quantization based \herd, while for \herdsm, we use \herdlrq. Note that the  combination of \herdsm + \herdffn is $1.47\times$ faster than baseline despite almost matching accuracy.}\vspace*{-4pt}
\label{tab:mlp-main-table}
\begin{center}
\begin{tabular}{|c|c|c|c|c|}
 \hline
 \multicolumn{2}{|c|}{} & Baseline & $\Mffn$ (HiRE-Q) & \makecell{$\Mffn$ (HiRE-Q) \\ +  $\Msm$ (HiRE-Q + HiRE-LR)} \\
 \hline
 \multirow{2}{*}{\makecell{Pre-training \\ Performance}} & Top1 Accuracy & 57.15\% & 57.03\% & 56.93\% \\
 \cline{2-5}
  & Perplexity & 2.045 & 2.056 & NA \\
 \hline
 \hline
 \multirow{4}{*}{\makecell{Downstream \\ Performance}} & Machine Translation & 47.92 & 46.95 &  46.94 \\ 
 \cline{2-5}
 & SuperGLUE Benchmark & 62.0 & 62.49 &  61.74 \\
 \cline{2-5}
 & Question Answering & 29.65 & 30.88 &  30.86 \\
 \cline{2-5}
 & Discriminative Tasks & 51.69 & 51.14 &  50.08 \\
 \hline
 \hline
 \multicolumn{2}{|c|}{\emph{Speedup}} & 1.0x & 1.16$\times$ & 1.47$\times$ \\
 \hline
\end{tabular}
\end{center}
\end{table*}

\begin{table*}[t!]
\vspace*{-4.0pt}
\caption{\distop (Algorithm~\ref{alg:distop}): Quality and latency evaluations for the \herdffn model deployed on a $2\times 2$ slice of Google Cloud TPUv5e. Clearly performing distributed, approximate top-$k$ computation is critical to obtain speedups during deployment on multiple machines, while not losing any quality on average.}\vspace*{-5pt}
\begin{center}
\begin{tabular}{|c|c|c|c|}
 \hline
 \multicolumn{2}{|c|}{} & \makecell{$\Mffn$ (HiRE-Q)} & \makecell{$\Mffn$ + HiRE-Q \\ + (\distop)} \\
 \hline
 \hline
 \multirow{2}{*}{\makecell{Pre-training \\ Performance}} & Top1 Accuracy & 57.03\% & 56.82\% \\
 \cline{2-4}
  & Perplexity & 2.056 & 2.064 \\
\hline
 \hline
 \multirow{4}{*}{\makecell{Downstream \\ Performance}} & Machine Translation & 46.95 & 47.03 \\ 
 \cline{2-4}
 & SuperGLUE Benchmark & 62.49 & 61.37 \\
 \cline{2-4}
 & Question Answering & 30.88 & 30.11 \\
 \cline{2-4}
 & Discriminative Tasks & 51.14 & 50.33 \\
 \hline
  \hline
 \multicolumn{2}{|c|}{\emph{Speedup}} & 1.0$\times$ & 2.27$\times$ \\
 \hline
\end{tabular}
\end{center}
\label{tab:approx-gather-24b-main-table}
\end{table*}

\subsection{Evaluation}
We evaluate the performance of different models both in terms of quality as well as inference latency. We measure quality using evaluation on pre-training dataset as well evaluations on multiple downstream datasets. Following standard works in the domain, we compute downstream performance of the model by applying it to multiple tasks, each focusing on a specific capability of LLMs. Collectively these datasets provide us a thorough assessment framework. To facilitate evaluation of our pretrained base models we perform 1-shot evaluations on all datasets.
\begin{enumerate}[leftmargin=*,noitemsep,nolistsep]
        \item \textbf{Machine Translation}: We use English-French tasks from WMT14 \cite{bojar2014findings} and German-French tasks from WMT22 \cite{zerva2022findings}.
        \item \textbf{SuperGLUE Benchmark}: We use multiple SuperGLUE \cite{wang2019superglue} tasks such as BoolQ \cite{clark2019boolq} etc. See appendix for the full list of datasets.
        \item \textbf{Question Answering}: We use SQuADv2 \cite{rajpurkar2018know}, TriviaQA \cite{joshi2017triviaqa}, NaturalQuestions \cite{kwiatkowski2019natural}, WebQuestions \cite{bordes2014question} and TyDiQA-English \cite{clark2020tydi}.
        \item \textbf{Discriminative Tasks}: We use additional tasks to evaluate \emph{commonsense reasoning}, \emph{ranking} and \emph{textual entailment}. See appendix for the full list of datasets.
\end{enumerate}
For Question Answering evaluations we use \textit{Exact Match} as our metric (Except \textit{F1-score} for TyDiQA), \textit{Accuracy} for Discriminative tasks and \textit{Character n-gram F-score} \cite{popovic2015chrf} for machine translation. For each task group, we report the Macro-Average of task metrics in our downstream evaluations.

For evaluating inference latency, we use a single TPUv5e device with batch size of one, and for generating one full response per query. We also demonstrate the importance of \distop in scaling to multiple devices, by presenting results for a slice of $4$ TPUv5e devices. We report \emph{Speedups} by measuring number of response tokens generated per unit time step.

\vspace{-3mm}
\subsection{Results}
{\bf \herdsm}: Table~\ref{tab:softmax-main-table} presents a comparison of $\Msm$ (i.e., \herd applied to the softmax layer of $\Mbase$) with the baseline dense model ($\Mbase$). Note that both \herdq and \herdlr provide about $1.16\times$ speedup over baseline while providing almost the same next-token prediction accuracy on the pretraining dataset. Downstream accuracy is also within $0.2\%$ of the baseline. We observe that combining $\herdq$ and $\herdlr$ improves the latency speedup to $1.22\times$ with less than $0.5\%$ overall drop in accuracy.

{\bf \herdffn}: Table~\ref{tab:mlp-main-table} presents a comparison of both \herdffn ($\Mffn$) as well as the full \herd approach applied to both the Softmax and FFN layers ($\Mfull$), with respect to the original model ($\Mbase$). Similar to softmax layer, here again combining \herdlr and \herdq leads to as much as $1.47\times$ latency reduction while maintaining almost similar pre-training and downstream evaluation accuracy. 

{\bf Multiple devices}: In latency critical applications, LLMs are served on multiple accelerator devices, with their weights distributed across them.
Implementing \herd as is on multiple devices in this fashion turns out to be suboptimal due to the cost of identifying top-$k$ activations jointly across all the devices. To overcome this, we proposed a distributed version of the top-$k'$ operation~\distop, which first performs top-$\frac{k'}{d}$ operation on each of the $d$ devices and then takes a union of those activations to approximate top-$k'$ (Algorithm~\ref{alg:distop}). As we can see from Table~\ref{tab:approx-gather-24b-main-table} improves the latency by $2.27\times$, compared to the vanilla implementation of~\herd (Algorithm~\ref{alg:herd}), with comparable quality on average across downstream tasks.

{\bf Why high recall estimation?}
\begin{table*}[t!]
\caption{Importance of High Recall for \herdsm. This table presents the pretraining top-$1$ accuracy as well as the size of intersection with the top-$k$ tokens of the original model, when we use different values of $k'$ in the approximate computation step and $k=32$. $k'=\textrm{None}$ refers to the setting where we do not follow the approximate computation by an exact computation. As we can see, (i) not doing the exact computation (i.e., $k'=\textrm{None}$) leads to large drop in top-$1$ accuracy. Similarly, using a $k'$ that is much larger than $k$ is critical in obtaining the top-$k$ tokens correctly.}\vspace*{-5pt}
\begin{center}
\begin{tabular}{|c|c|c|c|c|c|c|c|c|}
 \hline
 \multicolumn{2}{|c|}{\herdlr for Softmax (r=25\%)} & $k'=\textrm{None}$ & $k'=32$ & $k'=64$ & $k'=128$ & $k'=256$ & $k'=384$ & $k'=512$ \\
 \hline
 \multirow{2}{*}{\makecell{Pre-training \\ Performance}} & Top1 Accuracy & 51.87\% & 56.43\% & 56.63\% & 56.77\% & 56.86\% & 56.89\% & 56.92\% \\
 \cline{2-9}
  & Top32 Intersection & 19.24 & 19.24 & 24.13 & 26.94 & 28.72 & 29.26 & 29.93 \\
 \hline

 \multicolumn{9}{c}{} \\
 
 \hline
 \multicolumn{2}{|c|}{\herdq for Softmax (\intfour)} & $k'=\textrm{None}$ & $k'=2$ & $k'=4$ & $k'=8$ & $k'=32$ & $k'=64$ & $k'=128$ \\
 \hline
 \multirow{2}{*}{\makecell{Pre-training \\ Performance}} & Top1 Accuracy & 55.94\% & 56.76\% & 56.94\% & 56.98\% & 56.99\% & 56.99\% & 56.99\% \\
 \cline{2-9}
  & Top32 Intersection & 27.68 & 1.98 & 3.96 & 7.90 & 27.68 & 30.86 & 31.30\\
\hline
\end{tabular}
\end{center}
\label{tab:ablation_high_recall}
\end{table*}

Consider Algorithm \ref{alg:herd} and \eqref{eqn:general-topk}. We define Recall as the intersection between the set of elements, $S'$ computed using $\Zapprox$ matrix and $S$ using $\Z$ matrix. In case of Softmax, we compute the intersection between $S'$ and $S$ for both \herdlr and \herdq in Table \ref{tab:ablation_high_recall}. We see a systematic increase of recall with increase in $k'$ of the algorithm.
In case of $\Mffn$, we evaluate models with different ($k'$, $k$) pairs to study the importance of high \emph{recall} in pretraining metrics.
From Table~\ref{tab:high-recall-mlp}, we observe that, as we increase $k'$, the model shows an improvement, further signifying the importance of high recall.

{\bf Importance of learned projections in \herdlr}: While \herdq is much easier to implement since it does not require any training, the potential for gains is limited due to the inherent nature of quantization. On the other hand, \herdlr has the potential to deliver larger latency improvements, but requires retraining as discussed in Section~\ref{sec:exp-training}. In this section, we explore whether random projections can be directly used instead of learning them. Table~\ref{tab:random-proj-vs-learned-proj} in Appendix~\ref{app:additional-results} demonstrates that random projections can lead to as much as $10\%$ reduction in downstream evaluation accuracy at similar rank justifying our approach of learned low-rank projections.

{\bf Location of sparse layers}: In all the results so far, we have modified only the odd FFN layers to be sparse (\eqref{eqn:herd-groupsparse}), while the remaining were dense (\eqref{eqn:feedforward}). In Table~\ref{tab:architecture}, we present results for different choices of sparse layers, which demonstrate that modifying all the layers to be sparse leads to substantial drop in accuracy, and the location of sparse layers makes an effect on the overall accuracy. It is interesting to identify the optimal selection of sparse layers.

\begin{table}[h]
\caption{Commonpath Experimental Results: Static sparsity denotes the percentage of feedforward neurons used by all tokens i.e. commonpath while Adaptive Sparsity denotes the percentage of neurons activated by feedforward neurons not in commonpath.}
\begin{center}
\resizebox{\columnwidth}{!}{
\begin{tabular}{|c|c|c|c|c|}
 \hline
 \multicolumn{2}{|c|}{} & Baseline &  Sparse &  CommonPath \\
 \hline
 \multirow{3}{*}{\makecell{Sparsity}} & Static Sparsity & 100\% & 0\% &  16.66\%  \\
  \cline{2-5}
  & Adaptive Sparsity & NA & 5.55\% &  3.33\%  \\
 \hline
  \hline
 \multirow{2}{*}{\makecell{Pre-training \\ Performance}}  & Top1 Accuracy & 57.15\% & 57.03\% &  57.03\%  \\
 \cline{2-5}
 & Perplexity & 2.045 & 2.056 &  2.055  \\
 \hline
 \hline
 \multirow{4}{*}{\makecell{Downstream \\ Performance}} & MT & 47.92 & 46.95 &  46.93 \\ 
 \cline{2-5}
 & SuperGLUE  & 62.07 & 62.49 &  61.66 \\
 \cline{2-5}
 & Q \& A & 29.65 & 30.88 &  29.94 \\
 \cline{2-5}
 & Disc. Tasks & 51.69 & 51.14 &  51.14 \\
 \hline
 \hline
 {} & \emph{Speedup} & 1.0$\times$  & 1.16 $\times$ &  1.22$\times$ \\
\hline
\end{tabular}}
\end{center}
\label{tab:commonpath}
\end{table}

{\bf Common path for increasing sparsity further}:
We now present results demonstrating the utility of a common path, to further enhance activation sparsity as described in Section~\ref{sec:overlap},~\eqref{eqn:commonpath-ffn}. The results, presented in Table~\ref{tab:commonpath} show that our commonpath technique is indeed able to improve dynamic sparsity by $2\%$ while still maintaining accuracy, which leads to about $6\%$ lower latency.  

\begin{figure}[t!]
\centering
\includegraphics[width=0.5\textwidth]{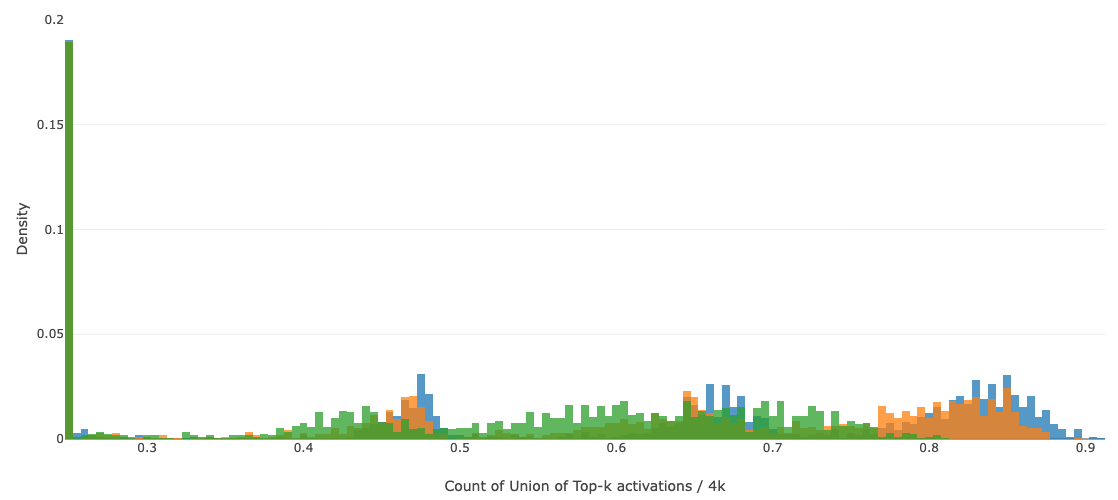}
\caption{Dynamic overlap of top-$k$ activations across related responses while generating $4$ parallel samples for the same query. On x-axis is the size of union of top-$k$ activations across $4$ generations divided by $4k$, for $k=(0.05)*m$. As we can see, there is substantial fraction of mass away from $1$, suggesting that the top-$k$ activations of related responses have high overlap, which can yield further latency improvements with \herd.}
\label{fig:adaptive-sparsity-plot}
\end{figure}

 {\bf Exploiting activation overlap for larger num samples with adaptive sparsity}: 
It has been shown that generating multiple diverse responses for a single query, and subsequent ranking aids in selecting the most suitable output~\cite{mudgal2023controlled}. We hypothesize that such responses share significant overlap in their non-zero neural activations, offering potential for enhancing effective sparsity in FFN layers. This is supported empirically (Figure~\ref{fig:adaptive-sparsity-plot}), where the union of non-zero activations across multiple responses is consistently smaller than the sum of number of individual response activations.


\section*{Broader Impact Statement}
We expect utilizing \herd can lead to more energy efficient large model inference. Beyond that there are potential societal consequences of helping large models become more accessible, none which we feel must be specifically highlighted here. 
\bibliography{main_icml}
\bibliographystyle{icml2024}

\newpage
\appendix
\onecolumn

\section{Dataset details}
\begin{enumerate}[leftmargin=*,noitemsep,nolistsep]
        \item \textbf{Machine Translation}: We use English-French tasks from WMT14 \cite{bojar2014findings} and German-French tasks from WMT22 \cite{zerva2022findings}
        \item \textbf{SuperGLUE Benchmark}: We use multiple SuperGLUE \cite{wang2019superglue} tasks such as BoolQ \cite{clark2019boolq}, CommitmentBank \cite{de2019commitmentbank}, Choice of Plausible Alternatives \cite{roemmele2011choice}, Multi Sentence Reading Comprehension \cite{khashabi2018looking}, Recognizing Textual Entailment, Words in Context \cite{pilehvar2018wic}, Winograd Schema Challenge \cite{levesque2012winograd} and Reading Comprehension with Commonsense Reasoning (ReCoRD) \cite{zhang2018record}.
        \item \textbf{Question Answering}: We use SQuADv2 \cite{rajpurkar2018know}, TriviaQA \cite{joshi2017triviaqa}, NaturalQuestions \cite{kwiatkowski2019natural}, WebQuestions \cite{bordes2014question} and TyDiQA-English \cite{clark2020tydi}.
        \item \textbf{Discriminative Tasks}: We use additional tasks to evaluate \emph{commonsense reasoning} (ARC \cite{clark2018think}, HellaSwag \cite{zellers2019hellaswag}, WinoGrande \cite{sakaguchi2021winogrande}, PIQA \cite{bisk2020piqa} and Story Cloze \cite{mostafazadeh2016corpus}), \emph{question answering} (RACE, \cite{lai2017race} and OpenBookQA \cite{mihaylov2018can}) and \emph{Textual Entailment} (ANLI \cite{nie2019adversarial}).
\end{enumerate}

\section{Additional Results}\label{app:additional-results}
\begin{figure}[h]
\centering
\includegraphics[width=0.95\textwidth]{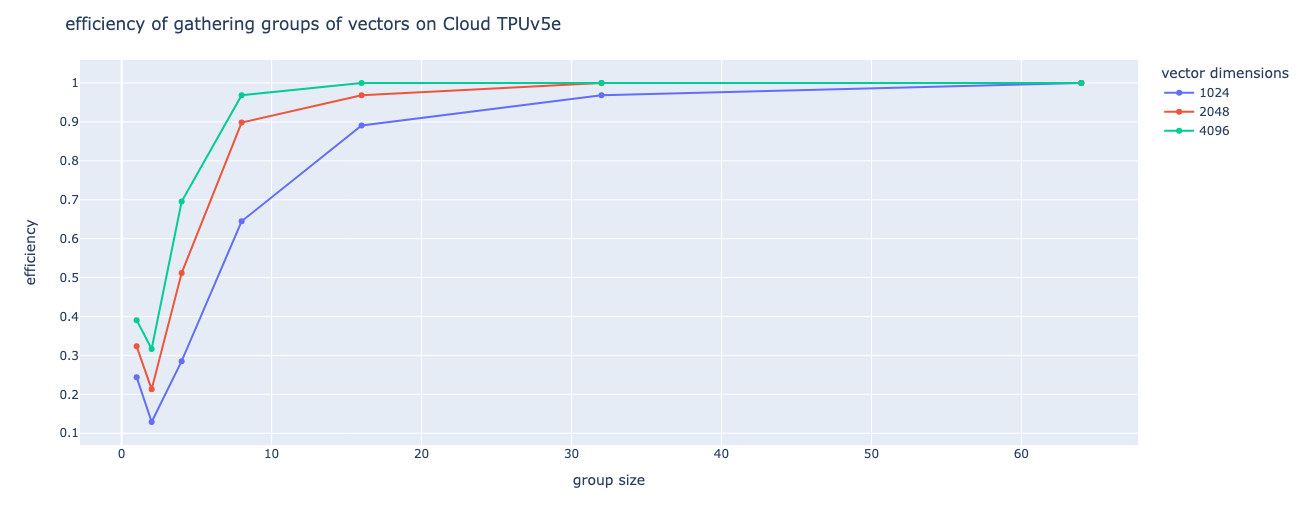}
\caption{Efficiency of memory transfer vs group sizes: For a tensor of dimension $n \times g \times d$, which we consider as $n$ groups, each with $g$ vectors of dimension $d$, we plot the efficiency of transferring a random (non-contiguous) subset of groups from HBM to cache as we vary the group size $g$ on the x-axis. Efficiency is defined as the time taken by the sparse operation divided by the time taken by an equivalent dense operation moving the same number of bytes. The numbers are computed for Cloud TPUv5e. As is clear from the figure, even small group sizes such as $8$ lead to very high efficiency, motivating the group sparse structure in our feedforward layers.}
\label{fig:group-sparsity-efficiency}
\end{figure}

\begin{table*}[h]
\caption{Do we need to recall more elements than $k$ in \herdffn?: Each column contains pretraining metrics for the \herdffn model by using various values of $k'$ and $k$ (indicated in the table as $(k',k)$), in comparison with the dense baseline model $\Mbase$. \textit{Full} denotes $k'$=$m$ where $m$ is the total number of hidden units. Baseline refers to dense model trained without the top-$k$ operation. As we can see from the results, using a larger $k'$ compared to $k$ is critical for bridging the quality gap between the dense and top-$k$ models.}
\begin{center}
\begin{tabular}{|c|c|c|c|c|c|c|}
 \hline
 \multicolumn{2}{|c|}{}  & (128, 128) & (192, 128) & (256, 128) & (Full, 128) & Baseline \\
 \hline
 \multirow{2}{*}{\makecell{Pre-training \\ Performance}} & Top1 Accuracy   & 56.81\% & 56.96\% & 57.03\% & 57.05\% & 57.15\% \\
 \cline{2-7}
  & Perplexity  & 2.071 & 2.06 & 2.056 & 2.054 & 2.045 \\
\hline
\end{tabular}
\end{center}
\label{tab:high-recall-mlp}
\end{table*}

\begin{table*}[h]
\caption{Importance of learned low rank matrix. This table presents the pretraining metrics by using random vs learned low rank projection. As we can see, learning the projection matrix is crucial for maintaining quality.}
\begin{center}
\begin{tabular}{|c|c|c|c|c|c|c|c|}
 \hline
 \multicolumn{2}{|c|}{} & \multirow{2}{*}{Baseline} & Trained, k=384 & \multicolumn{4}{c|}{Random, k=384} \\
 \cline{4-8}
 \multicolumn{2}{|c|}{} & & r=25\% & r=25\% & r=33\% & r=50\% & r=66\% \\
 \hline
 \multirow{2}{*}{\makecell{Pre-training \\ Performance}} & Top1 Accuracy & 57.40\% & 57.29\% & 46.78\% & 49.26\% & 54.02\%  & 55.81\% \\
 \cline{2-8}
  & Top32 Intersection & 32.0 & 29.26 & 15.53 & 18.42 & 24.21 & 26.89 \\
\hline
\end{tabular}
\end{center}
\label{tab:random-proj-vs-learned-proj}
\end{table*}

\begin{table*}[h]
\caption{Ablation on architectures: We train models with multiple sparse FFN layer configurations and evaluate with approximate $k'$ = 256 and $k$ = 128 for all models. $L$, $D$ and $S$ denote the number of layers in the model, dense layer (\eqref{eqn:feedforward}) and sparse layer (\eqref{eqn:herd-groupsparse}) respectively. $(D,S)$ means alternating dense and sparse layers, while Last$L/2$ refers to the final $L/2$ layers being sparse and the remaining being dense.}
\begin{center}
\begin{tabular}{|c|c|c|c|c|c|c|c|}
 \hline
 \multicolumn{2}{|c|}{} & Baseline & ($D$ $S$) x $L/2$ & Last $L/2$ & ($D$ $S$ $S$) x $L/3$ & Last $2L/3$ & $L$ layers \\
 \hline
 \multirow{2}{*}{\makecell{Pre-training \\ Performance}} & Top1 Accuracy & 57.15\% & 57.03\% & 57.03\% & 56.84\% & 56.83\% & 56.24\% \\
 \cline{2-8}
  & Perplexity & 2.045 & 2.056 & 2.058 & 2.069 & 2.071 & 2.108 \\
\hline
 \hline
 \multirow{4}{*}{\makecell{Downstream \\ Performance}} & Machine Translation & 47.92 & 46.95 & 46.35 & 46.17 & 45.83 & 44.35\\ 
 \cline{2-8}
 & SuperGLUE Benchmark & 62.07 & 62.49  & 60.43 & 59.37 & 60.74 & 59.3 \\
 \cline{2-8}
 & Question Answering & 29.65 & 30.88 & 29.86 & 28.88 & 29.09 & 27.51 \\
 \cline{2-8}
 & Discriminative Tasks & 51.69 & 51.14 & 50.99 & 50.37 & 50.8 & 49.65 \\
 \hline
  \hline
 {} & \emph{Speedup} & 1.0x & 1.16x & 1.18x & 1.17x & 1.18x & 1.44x \\
 \hline
\end{tabular}
\end{center}
\label{tab:architecture}
\end{table*}




\end{document}